\documentclass[12pt]{article}

\usepackage[utf8]{inputenc}
\usepackage{url}
\usepackage{soul}
\usepackage[T1]{fontenc}
\usepackage[colorinlistoftodos]{todonotes}
\usepackage{graphicx}
\usepackage{epstopdf}
\usepackage[justification=raggedright]{caption}
\usepackage{subcaption}
\usepackage{graphics}
\usepackage{placeins}
\usepackage{rotating}
\usepackage[margin=0.985in]{geometry}
\usepackage{threeparttable} 
\usepackage[round]{natbib}
\usepackage{amsmath,amssymb}
\usepackage{setspace}
\usepackage{listings}
\usepackage{graphicx}

\def\refname{Further reading}

\providecommand{\keywords}[1]
{
  \small	
  \textbf{\textit{Keywords---}} #1
}

\title{\textbf{Machine learning in policy evaluation:
new tools for causal inference}}
\author{No\'emi Kreif and Karla DiazOrdaz }

\date{}

\begin{document}

\maketitle

\begin{abstract}
\begin{singlespace}

While machine learning (ML) methods have received a lot of attention in recent years, these methods are primarily for prediction. Empirical researchers conducting policy evaluations are, on the other hand, pre-occupied with causal problems, trying to answer counterfactual questions: what would have happened in the absence of a policy? Because these counterfactuals can never be directly observed (described as the ``fundamental problem of causal inference'') prediction tools from the ML literature cannot be readily used for causal inference. In the last decade, major innovations have taken place incorporating supervised ML tools into estimators for causal parameters such as the average treatment effect (ATE). This holds the promise of attenuating model misspecification issues, and increasing of transparency in model selection.   
One particularly mature strand of the literature include approaches that incorporate supervised ML approaches in the estimation of the ATE of a binary treatment, under the \textit{unconfoundedness}  and positivity assumptions (also known as exchangeability and overlap assumptions). 

This article begins by reviewing popular supervised machine learning algorithms, including trees-based methods and the lasso, as well as ensembles, with a focus on the Super Learner. Then, some specific uses of machine learning for treatment effect estimation are introduced and illustrated, namely (1) to create balance among treated and control groups, (2) to estimate so-called nuisance models (e.g. the propensity score, or conditional expectations of the outcome) in semi-parametric estimators that target causal parameters (e.g. targeted maximum likelihood estimation or the double ML estimator), and (3) the use of machine learning for variable selection in situations with a high number of covariates. 

Since there is no universal best estimator, whether parametric or data-adaptive, it is best practice to incorporate a semi-automated approach than can select the models best supported by the observed data, thus attenuating the reliance on subjective choices.

  \end{singlespace}  
    
\end{abstract}

\keywords{Machine learning, causal inference, treatment effects, health economics, program evaluation, policy evaluation, doubly robust methods, matching}

\begin{flushleft}

\section{Overview}

Most scientific questions, such as those asked when evaluating policies, are causal in nature, even if they are not specifically framed as such. Causal inference reasoning helps clarify the scientific question, and define the corresponding causal estimand, i.e the quantity  of interest, such as the average treatment effect (ATE). It also makes clear the assumptions necessary to express the estimand in terms of the observed data, known as identification. Once this is achieved, the focus shifts to estimation and inference. While machine learning methods have received a lot of attention in recent years, 
these methods are primarily geared for prediction. There are many excellent texts covering machine learning focussed on prediction \citep{friedman2001elements,james2013introduction}, but not dealing with causal problems. Recently, some authors within the Economics community have started examining the usefulness of machine learning for the causal questions that are typically the subject of applied econometric research \citep{varian2014big,kleinberg2015prediction,mullainathan2017machine, athey2017impact,athey2017beyond,athey2017state}.

In this Chapter, we contribute to this literature by providing an overview and an illustration of machine learning methods for causal inference, with a view to answer typical causal questions in policy evaluation, and show how these can be implemented with widely used statistical packages. We draw on innovations form a wide range of quantitative social and health sciences, including economics, biostatistics and political science.

We focus on methods to estimate the ATE of a binary treatment (or exposure), under ``no unobserved confounding assumptions'' (see Section \ref{estimands}). 
The remainder of this article is as follows. First, we introduce the notation and the assumptions for the identification of causal effects. Then we outline our illustrative treatment effect estimation problem,  an impact evaluation of a social health insurance program in Indonesia.
Next, we provide a brief introduction to supervised machine learning methods.
In the following sections, we review methods for estimating the ATE. These can be roughly categorised into three main types: methods that aim to balance covariate distributions  (propensity scores (PS) and other matching methods), methods that fit outcome regressions to ``impute'' potential outcomes and estimate causal effects, and the so called ``double robust'' methods that combine these. We also discuss the use of machine learning for variable selection, a challenge increasingly important with ``Big data'', especially with a large number of variables. 
In the last section we provide a brief overview of developments for other settings and a discussion. 

\section{Estimands and assumptions for identification}\label{estimands}
\subsection{Notation and assumptions}
Let $A$ be an indicator variable for treatment, and $Y$ be the outcome of interest. Denote by $Y^a_i$ the \emph{potential outcome} that would manifest if the $i$-th subject were exposed to level $a$ of the treatment, with $a\in \{0, 1\}$. The observed outcome can then be written as $Y_i=Y^0_i(1-A_i) + Y^1_i A_i$    \citep{rubin1978bayesian}. 

Throughout, we assume the  Stable Unit Treatment Value Assumption (SUTVA) holds,
which comprises  \textit{no interference}, i.e. the potential outcomes of the $i$-th individual are unrelated to the treatment status of all other individuals, and \textit{consistency}, i.e. for all individuals $i=1,\ldots, N$, if $A_i=a$ then $Y_i^a=Y_i$, for all $a$ \citep{robins2000marginal,VanderWeele2009consistency,cole2009consistency,pearl2010consistency}.  

Denote the observed data of each individual by  $O_i=(\mathbf{X}_i,A_i,Y_i)$, where $\mathbf{X}_i$ is a vector of  confounding variables, that is factors that influence simultaneously the potential outcomes and treatment. We assume that the data are an independent identically distributed sample of size $N$.

Individual level causal effects are defined as the difference between these ``potential outcomes''. Researchers are often interested in the average of these individual causal effects over some population. A widely considered casual estimand is the ATE, defined as $\psi=E[Y_i^1-Y_i^0]$. Further estimands include the average taken over the treated subjects (the average treatment effect on the treated, ATT) or the conditional average treatment effect (CATE), which takes the expectation over individuals with certain observed characteristics. Here we focus on the ATE.

Since the potential outcomes  can never be simultaneously directly observed, these estimands cannot be expressed in terms of observed data, or \emph{identified}, without further assumptions.
 A commonly invoked assumption which we will make throughout is \emph{ignorability} or \emph{unconfoundedness} of the treatment assignment, (also known as conditional exchangeability). This assumption  requires that the potential outcomes are independent of treatment, conditional on the observed covariates,
\begin{equation}\label{unconf}
A_i \perp (Y^0_i,Y^1_i) | \mathbf{X}_i=\mathbf{x}.
\end{equation}

The plausibility of this assumption needs to be carefully argued in each case, ideally with careful data collection and based on subject matter knowledge about the variables that may be associated with the outcome as well as influencing the treatment, as it cannot be tested using the observed data \citep{rubin2005causal}. 

The second necessary assumption is the \emph{positivity} of the treatment assignment (also referred as ``overlap''): \begin{equation} \label{pos} 0<P(A_i=1|\mathbf{X}_i=\mathbf{x})<1,\end{equation}
implying that for any combination of covariates, there is a nonzero probability of receiving both the treatment and control states.

Using the unconfoundedness and the positivity assumptions, the conditional mean of the potential outcomes corresponds with the conditional mean of the observed outcomes, $E[Y_i^1|\mathbf{X}_i,A_i=1]=E[Y_i|\mathbf{X}_i,A_i=1]$ and $E[Y_i^0|\mathbf{X}_i,A_i=0]=E[Y_i|\mathbf{X}_i,A_i=0]$, and the ATE can be identified by:
\begin{equation}\label{ATE}
\psi = E[Y_i|\mathbf{X}_i,A_i=1] - E[Y_i|\mathbf{X}_i,A_i=0].
\end{equation}

\section{Illustrative example: the impact of health insurance on assisted birth in Indonesia}\label{Sec:Birthdata}

We illustrate the methods by applying them each in turn to an impact evaluation of a national health insurance programme in Indonesia (more details in \citet{kreif2018evaluating}). The dataset consists of births between 2002 and 2014, extracted from the Indonesian Family Life Survey (IFLS). The policy of interest, i.e. the treatment, is ``being covered by the health insurance offered for those in formal employment and their families'' (contributory health insurance). We are interested in the ATE of such health insurance on the probability of the birth being assisted by a health care professional (physician or midwife). We construct a list of observed covariates including the mother's characteristics, (age, education, wealth in quintiles) and household's characteristics (social assistance, experienced a natural disaster, rurality, availability of health services: a village midwife, birth clinic, hospital).

We expect that the variables describing socioeconomic status may be particularly important, because those with contributory insurance tend to work in the formal sector, and have higher education than those uninsured, and these characteristics would make a mother more likely to use health services even in the absence of health insurance. Similarly, the availability of health services is expected to be an important confounder, as those who have health insurance may live in areas where it is easier to access health care, with or without health insurance. 

The final dataset reflects typical characteristics of survey data: the majority of variables are binary, with two variables categorical and one continuous (altogether 34 variables). Due to the nature of the survey, for around one third of women we could not measure confounder information from the past, but had to impute it with information from the time of the survey. Two binary variables indicate imputed observations. 

  
For simplicity, any records with any other missing data have been list-wise deleted.  This approach provides unbiased estimates of the ATE as long as missingness does not depend on both the treatment and the outcome \citep{Bartlett2015}.
The resulting complete-case dataset consists of 10985 births, of whcih 1181 are in the treated group, as the mother had health insurance in the year of the child's birth, while 8574 babies had their birth assisted by a health professional.

\section{Introduction to machine learning for causal inference}\label{Sec:MLIntro}
\subsection{Supervised machine learning}
The type of machine learning tools most useful for causal inference are those labelled as ``supervised machine learning''. These tools, similarly to regression, can summarise linear and non-linear relationships in the data  and  can predict some  $Y$ variable given new values of covariates $(A,X)$ \citep{varian2014big}. A ``good'' prediction is defined in relation to a loss function,  for example the mean sum of squared errors. A commonly used measure of this is  the \emph{test mean squared error} (test MSE), defined as the average square prediction error among observations not previously seen. This quantity differs from the usual MSE calculated among observations that were used to fit the model. 
In the absence of a very large dataset that can be used to directly estimate the test MSE, it can be estimated by holding out a subset of the observation from the model fitting process, using the so-called ``V-fold cross-validation'' procedure (see e.g. \citet{zhang1993model}). When performing $V$-fold cross-validation, the researcher randomly divides the set of observations into $V$ groups (folds). The first group is withheld from the fitting process, and thus referred to as the \emph{test data}. The algorithm is then fitted using the data in the remaining $V-1$ folds, called the \emph{training data}. Finally, the MSE is calculated using the test data, thus evaluating the performance of the algorithm. This process is repeated for each fold, resulting in $V$ estimates of the test MSE, which are then averaged to obtain the so-called cross-validated MSE.  In principle, it is possible to perform cross-validation with just one split of the sample, though results are highly dependent on the sample split.  Thus, typically, $V=5$ or $V=10$ is used in practice.

The ultimate goal of machine learning algorithms is to get good out-of-sample predictions minimising the test MSE \citep{varian2014big}, typically achieved by a combination of flexibility and simplicity, often described as the ``bias-variance trade off''. The formula for the MSE shows why minimising it achieves this: $\mbox{MSE}(\hat {\theta })=\mbox{Var}({\hat{\theta}})+\mbox{Bias}({\hat {\theta}},\theta)^{2}$.  For example, a nonlinear model with many higher order terms is more likely to fit the data better than a simpler model, however it is unlikely that it will fit a new dataset similarly well, often referred to as ``overfitting''.   

\emph{Regularisation} is a general approach that aims to achieve balance between flexibility and complexity, by penalising more complex models. 
 With less regularisation, one does a better job at approximating the within-sample variation, but for this very reason, the  out-of-sample fit will typically get worse, increasing the bias. 
 The key question is choosing the level of regularisation: how to tune the algorithm so that it does not over-smooth or overfit.

In the context of selecting regularisation parameters, cross-validation can be used to perform so-called \emph{empirical tuning} to find the optimal level of complexity, where complexity  is often  indexed by the tuning parameters. The potential range of tuning parameters is divided into a grid (e.g. 10 possible values), and $V$ fold cross validation process is performed for each parameter value, enabling the researcher to choose the value of the tuning parameter with the lowest test MSE. Finally, the algorithm with the selected tuning parameter is re-fitted using all observations.     




\subsection{Prediction vs. causal inference}
 Machine learning is naturally suited to prediction problems, which have been traditionally treated as distinct from causal questions  \citep{mullainathan2017machine}. It may be tempting to interpret causally the output of the machine learning predictions, however making inferences from machine learning models is complicated by (1) the lack of interpretable coefficients for some of the algorithms, and (2) the lack of standard errors \citep{athey2017beyond}.
Moreover, for certain ``regression-like'' algorithms (e.g. Lasso), selecting the best model using cross validation, and then doing inference for the model parameters, ignoring the selection process, though common in practice, should be avoided as it leads to potential biases stemming from the shrunk coefficients, and underestimation of the true variance in the parameter estimates \citep{mullainathan2017machine}.

The causal inference literature (see e.g. \citet{Kennedy2016,van2011targeted,petersen2014applying}) stresses the importance of first defining the causal estimand of interest (also referred to as `'target parameter''), and then carefully thinking about the necessary assumptions for identification. 
Once the causal estimand has been mapped to an estimator (a functional of the observed data), via the identification assumptions, the problem becomes an estimation exercise. 
In practice many estimators involve models for parameters (e.g. conditional distributions, means), which are not of interest per se, but are necessary  to estimate the target parameter, these are called \emph{nuisance} models. Nuisance models estimation  can be thought of as  prediction problem  for which machine learning can be used. Examples include the estimation of propensity scores, or outcome regressions that can later be used to predict potential outcomes (see Section \ref{reg}). So while most machine learning methods cannot be readily used to infer causal effects, they can help the process.

A potential key advantage of using machine learning for the nuisance models is that it fits and compares many alternative algorithms, by for example, using cross validation (while cross-validation can be used to select among parametric models as well). Selecting models based on a well defined loss functions (e.g. the cross-validated MSE) can, beyond improving model fit, benefit the overall transparency of the research process \citep{athey2017impact}. 
 This is in  contrasts with how model selection is usually viewed in Economics, where model is chosen based on theory and estimated only once.

This has led to many researchers using machine-learning for the estimation of the nuisance parameters of standard estimators (e.g., outcome regression, inverse probability weighting by the propensity score, see e.g. \citet{lee2010improving, westreich2010propensity}). However, the behavior of these estimators is can be poor, resulting in slower convergence rates
and confidence intervals which are difficult to construct \citep{VanderVaart2014}. In addition, the resulting estimators are irregular and the nonparametric bootstrap is in general not  valid \citep{Bickel1997}.


An increasingly popular strategy to avoid these biases and have valid inference is to use the so-called doubly robust estimators (combining nuisance models for the outcome regressions and the propensity score) which we review in Section \ref{Sec:DR}. This is because DR estimators can converge at fast rates ($\sqrt{N}$) to the true parameter, and are therefore consistent asymptotically normal, even when the nuisance models have been  estimated via machine learning.

In the following sections, we briefly describe the machine learning approaches that have been most widely used for nuisance model prediction in causal inference, either because of their similarity to traditional regression approaches, their easy implementation due to the availability of statistical packages, their superior performance in prediction, or a combination of these.

\subsection{Lasso}\label{Sec:lasso}

LASSO (Least Absolute Shrinkage and Selection Operator)
is a penalised linear (or generalised linear) regression algorithm, fitting the  model including all $d$ predictors. It aims to find the set of coefficients that minimise  the  sum-of-squares loss function, but subject to a constraint on the sum of absolute values (or $\ell_1$ norm) of coefficients being equal to a constant $c$ often referred to as \emph{budget}, i.e.  $\sum^d_{j=1}\|\beta_j\|_{1}=c$.
This results in a (generalised) linear regression in which only a small number of covariates have nonzero coefficient: this absolute-value  regulariser induces a sparse coefficient vector. The nonzero coefficient estimates are also shrunk towards zero. This significantly reduces their variance at the ``price'' of increasing the bias.
An equivalent formulation of the lasso is 
 \begin{equation}
 \min _{\beta \in \mathbb {R} ^{d}}
 \left\{\left\| y-X \beta \right\|^{2}+\lambda \|\beta \|_{1}\right\},
 \end{equation}
with the penalty $\lambda$ being the  tuning parameter.

 As  $\lambda$ increases, the flexibility of the lasso regression fit decreases, leading to decreased variance but increased bias. Beyond  a certain point however, the decrease in variance due to
increasing  $\lambda$ slows, and the shrinkage on the coefficients causes them to be
significantly underestimated, resulting in a large increase in the bias. Thus the choice of $\lambda$  is critical. This is usually done by cross-validation, implemented by several R packages, e.g. \texttt{glmnet} and \texttt{caret}.

Because the lasso results in some of the coefficients being exactly  zero when the penalty $\lambda$ is sufficiently large, it essentially  performs variable selection.
The variable  selection however is driven by the tuning parameter, and it can happen that  some variables are selected in some of the CV 
partitions, but may be unused in another.  This problem is common when the variables are correlated with each other, and they explain very similar ``portions'' of the outcome variability. 
 A practical implication of this is that the researcher should remove from the set of potential variables those that are irrelevant, in the sense that they are very correlated to a combination of other, more relevant ones.

Another problem is inference after model selection, with some results \citep{Leeb2008} showing its is not possible to obtain  (uniform)
model selection consistency. As we demonstrate in Section \ref{Sec:VarSel} some uses of lasso enable consistent estimation post-variable selection.
 
\subsection{Tree based methods}

\subsubsection{Regression trees}
Tree based methods, also known as  classification and regression trees or ``CARTs'', have a similar logic to decision trees familiar to economists, but here the ``decision''  is a choice about how to classify the observation. The goal is to construct (or ``grow'') a decision tree that leads to good out-of-sample predictions.   They can be used for classification with binary or multicategory outcomes (``classification trees'') or with continuous outcomes (``regression trees''). A regression tree uses a recursive algorithm to estimate a function describing the relationship between a multivariate set of independent variables and a single dependent variable, such as treatment assignment.

Trees tend to work well for settings with nonlinearities and interactions in the outcome-covariate relationship. In these cases, they can improve upon traditional classification algorithms such  logistic regression. In order to avoid overfitting, trees are \emph{pruned} by applying tuning parameters that penalise complexity (the number of leaves). A major challenge with tree methods is that they  are sensitive to the initial split of the data, leading to high variance. Hence, single trees are rarely used in practice, but instead ensembles - algorithms that stack or add together different algorithms - of trees are used, such as random forest or boosted CARTs. 

\subsubsection{Random forests}
Random forests are constructed using bootsrapped samples of the data, and growing a tree where only a (random) subset of covariates is used for creating the splits (and thus the leaves). These trees are then averaged, which leads to a reduction in  variance.  The tuning parameters, which can be set or selected using cross validation, include the number of trees, depth of each tree, and the number of covariates to be randomly selected (usually recommended to be approximately $\sqrt{d}$ where $d$ is the number of available independent variables). Popular implementations include the R packages \texttt{caret} and \texttt{ranger}.

\subsubsection{Boosting}
Boosting generates a sequence of  trees where the first tree's residuals are used as outcomes for the construction of the next tree. Generalised boosted models add together many simple functions to estimate a smooth function of a large number of covariates. Each individual simple function lacks smoothness and is a poor approximation to the function of interest, but added together they can approximate a smooth function just like a sequence of line segments can approximate a smooth curve. In the implementation in the R package \texttt{gbm} \citep{mccaffrey2004propensity},  each simple function is a regression tree with limited depth. Another popular package is \texttt{xgboost}.

\subsubsection{Bayesian Additive Regression Trees}

Bayesian Additive Regression Trees (BARTs) can be distinguished from other tree based ensembling algorithms due to its underlying probability model \citep{kapelner2013bartmachine}. As a Bayesian model, BART consists of a set of priors for the structure and the leaf parameters and a likelihood for data in the terminal nodes. The aim of the priors is to provide regularisation, preventing any single regression tree from dominating the total fit.  To do this, BARTs employ so-called  ``Bayesian backfitting''  where the $j$-th tree is fit iteratively, holding all other $m-1$ trees constant by exposing only the residual response that remains unfitted. Over
many MCMC iterations, trees evolve to capture the fit left currently unexplained \citep{kapelner2013bartmachine}. BART is described as particularly well-suited to detecting interactions and discontinuities, and typically requires little parameter tuning \citep{hahn2017bayesian}. There is ample evidence on BART's good performance in predictions 
and even in causal inference \citep{hill2011bayesian,dorie2017automated}, and is implemented in several R packages (\texttt{bartMachine}, \texttt{dbarts}). Despite its excellent performance in practice,  there are limited theoretical results about BARTs.

\subsection{Super Learner ensembling}

\cite{varian2014big} highlights the importance of recognising uncertainty due to the model selection process, and the potential role ensembling can play in combining several models to create one that outperforms single models. Here we focus on the \emph{Super Learner} (SL) \citep{van2003unified}, a machine learning algorithm that uses cross validation to find the optimal weighted convex combination of multiple candidate prediction algorithms.  The algorithms pre-specified by the analyst form the \emph{library}, and can include parametric and machine learning approaches. The Super Learner has the \emph{oracle} property, i.e. it produces predictions that are at least as good as those of the best algorithm included in the library (see \citet{van2007super,van2011targeted} for details). 

Beyond its use for prediction \citep{polley2010super,rose2013mortality}, it has been used for PS and outcome model estimation (see for example, \citep{eliseeva2013application,van2014targeted,gruber2015ensemble}), and has been shown to reduce bias from model misspecification  \citep{porter2011relative, kreif2016evaluating,pirracchio2015improving}. Implementations of the Super Learner include the \texttt{SuperLearner}, \texttt{h2oEnsembleR} and the \texttt{subsemble} R packages, the latter two with increased computational speed to suit large datasets.

\section{Machine learning methods to create balance between covariate distributions}
\subsection{Propensity score methods}\label{PSmethods}

The propensity score (PS) \citep{rosenbaum1983central} defined as the conditional probability of treatment  $A$ given observed covariates, i.e. $p(\mathbf{x}_i)=P(A_i=1|\mathbf{X}_i=\mathbf{x}_i)$, is referred to as a ``balancing score'', due to its property of balancing the distributions of observed confounders amongst the treatment and control groups. The propensity score has been widely used to control for confounding, either for subclassification \citep{rosenbaum1984reducing}, as a metric to establish matched pairs in nearest neighbor matching \citep{rubin1996matching,abadie2016matching}, and for reweighting, using inverse probability of treatment weights \citep{hirano2003efficient}.  The latter two approaches have been demonstrated to have the best performance  \citep{lunceford2004stratification,austin2009some}.

The PS matching estimator constructs the missing potential outcome using the observed outcome of the closest observation(s) from the other group, and calculates the ATE as a simple mean difference between these predicted potential outcomes \citep{abadie2006large,abadie2011bias,abadie2016matching}. 

The \emph{inverse probability of treatment weighting} (IPTW) estimator for the ATE is simply a weighted mean difference between the observed outcomes of the treatment and control groups, where the weights $w_i$ are constructed from the estimated propensity score as 
\begin{equation}\label{weights}
w_i=\frac{A_i}{\hat{p}(\mathbf{X}_i)}+\frac{(1-A_i)}{1-\hat{p}(\mathbf{X}_i)}.
\end{equation}
With a correctly specified $p(\mathbf{X})$, $\psi^{IPTW}$ is consistent and efficient \citep{hirano2003efficient}.
The IPTW estimator can be expressed as  
\begin{equation}
\psi^{IPTW}= \frac{1}{N}\sum_{i=1}^N\frac{A_iY_i}{\hat{p}(\mathbf{X}_i)} - 
\frac{(1-A_i)Y_i}{1-\hat{p}(\mathbf{X}_i)}.
\end{equation}

Obtaining SEs for IPTW estimators can be done by the Delta method assuming the PS is known, or using robust covariance matrix, so-called sandwich estimator to acknowledge that the PS was estimated,  or by bootstrapping. IPTW estimators are sensitive to large weights.

 The validity of these methods depends on correctly specifying the PS model. In empirical work, typically probit or logistic regression models are used without interactions or higher order terms. However, the assumptions necessary for these to be correctly specified, for example the linearity of the relationship between covariates and probability of treatment in the logit scale, are rarely assessed \citep{westreich2010propensity}. More flexible modelling approaches, such as series regression estimation \citep{hirano2003efficient}, and  machine learning methods, including decision trees, neural networks and linear classifiers \citep{westreich2010propensity}, generalised boosting methods \citep{mccaffrey2004propensity,westreich2010propensity,lee2010improving,wyss2014role} or the Super Learner \citep{pirracchio2012evaluation} have been proposed to improve the specification of the PS.  However, even such methods may have poor properties, if their loss function targets measures of model fit (e.g. log likelihood, area under the curve) instead of balancing  covariates that are important to reduce bias \citep{westreich2011role}. 
 \citet{imai2014covariate} proposed a score that  explicitly balances the covariates, exploiting moment conditions that capture the desired mean independence between the treatment variable and the covariates that the balancing aims to achieve. A machine learning method for estimating propensity scores that aims to maximise balance is the  boosted CART approach \citep{mccaffrey2004propensity,lee2010improving}, implemented as the \texttt{TWANG} R package.  This approach  minimises a chosen loss function, based on covariate balance achieved in the IPTW weighted data, by iteratively forming a collection of simple regression tree models and adding them together to estimate the propensity score. It models directly the log-odds of treatment rather than  the propensity scores, to simplify computations. The algorithm can be specified to stop when the best balance is achieved. A recommended stopping rule is the  average standardised absolute mean difference (ASAM) in the covariates. 
 A balance metric,  the number of iterations, depth of interactions and shrinkage parameters need to be specified. The boosted CART approach to estimating PS has been demonstrated to improve balance and reduce bias in the estimated ATE \citep{lee2010improving,setoguchi2008evaluating} and has been extended to settings with continuous treatments \citep{zhu2015boosting}.

\subsection{Methods aiming to directly create balanced samples}
 
There is an extensive literature on methods that aim to create matched samples that are automatically balanced on the covariates, instead of  estimating and matching on a PS. 
An extension of Mahalanobis distance matching, the ``Genetic Matching'' algorithm \citep{diamond2013genetic}  searches a large space of potential matched treatment and control groups to minimise loss functions based on tests statistics  describing covariate imbalance (e.g. Kolmogorov-Smirnov tests). The accompanying \texttt{Matching} R package \citep{sekhon2011multivariate,mebane2011genetic} has a wide range of matching methods (including propensity score), matching options (e.g.  with or without replacement, 1:1 or 1:m matching),  estimands (ATE vs ATT) and balance statistics. 
The ``Genetic'' component of the  matching algorithm  chooses weights to give relative importance to the matching covariates to optimise the specified loss function.  The algorithm proposes batches of weights,  ``a generation'', and moves towards the batch of weights which maximise overall balance. Each generation is then used iteratively to produce a subsequent generation with better candidate weights. The ``population size'', i.e. the size of each generation is the tuning parameter to be specified by the user. 

 Similar approaches to creating optimal matched samples, with a different algorithmic solution are offered by \citet{zubizarreta2012using} and \citet{hainmueller2012entropy}. Both approaches use integer programming optimisation algorithms to construct comparison groups given balance constraints (maximum allowed imbalance) specified by the user, in the former case by one-to-many matching, in the latter case by constructing optimal weights. 

\subsection{Demonstration of balancing methods using the birth dataset}

We estimate a range of PS:  first, using a main terms logistic regression to estimate the conditional probability of being enrolled in health insurance, followed by two data-adaptive propensity scores. We include all covariates in the prediction algorithms, without  prior covariate selection. The first is a boosted CART, with  5000 trees, of a maximum depth of 2 interactions, and shrinkage of 0.005. The  loss function used is  ``average standardised difference''. 
\par
Second, we use the Super Learner with a library containing a range of increasingly data-adaptive prediction algorithms: 

\begin{itemize}
    \item 
logistic regression with and without all pair-wise interaction
\item generalised additive models with 2, 3 and 4 degrees of freedom,
   \item random forests -  including 4 random forest learners varying the number of trees (500, 2000), and the number of covariates to split on (5 and 8), implemented in the \texttt{ranger} R package),
   
   \item boosting - using the R package \texttt{xgboost}, with varying  number of trees (100 and 1000), shrinkage (0.001 and 0.1) and maximum tree depth (1 and 4)).
   \item a BART prediction algorithm using 200 regression trees with the tuning parameters set to default implementation in the \texttt{dbarts} R package.

\end{itemize}

We use 10-fold cross-validation and the mean sum of squares loss function. For the purposes of comparison, we have also implemented two 1:1 matching estimators with replacement, for the ATE parameter. First, we created a matched dataset based on the boosted CART propensity score, implemented without calipers. Second, we implemented the Genetic Matching algorithm, using a population size of 500, and a loss function that aims to minimise the largest p-values from paired t-test.   We have re-assessed the balance for the pair matched data. Throughout, we evaluate balance based on standardised mean differences, a metric that is comparable across weighting and matching methods \citep{austin2009some}. We calculate the ATE using IPTW and matching. 
The SEs for the IPTW are ``sandwich'' SEs while for the matching estimators  the Abadie-Imbens formula is used \cite{abadie2006large} that accounts for matching with replacement. 

Figure \ref{ps_risks} can be inspected to assess the relative performance of the candidate algorithms included in the SL. It displays the cross-validated estimates of the loss function (MSE) after an additional layer of cross-validation, so that the out-of sample performance of each individual algorithm and the convex combination of these algorithms (Super Learner) can be compared. The ``Discrete SL'' is defined as an algorithm that gives the best candidate the weight of 1. We see that the  convex Super Learner performs best. Table \ref{pscoefs} show the coefficients attributed to the different candidate algorithms in the final prediction algorithm that was used to estimate the PS. 

\begin{figure}
    \centering
    \caption{Estimated Mean Squared Error loss from candidate algorithms of the Super Learner}
    \label{ps_risks}
        \includegraphics[scale=0.85]{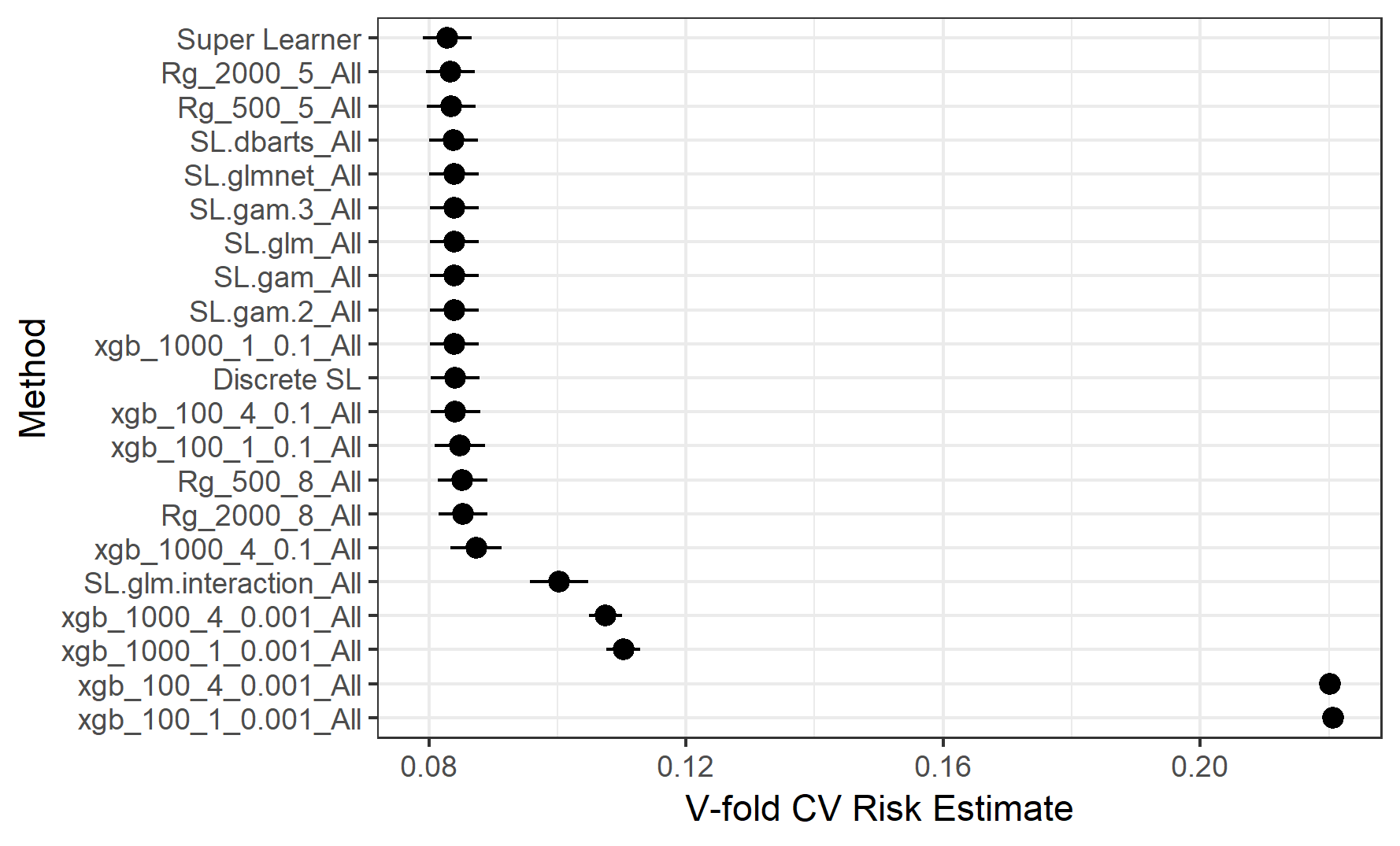}
        \begin{minipage}{0.65\textwidth}
    {\footnotesize Algorithms labelled by \texttt{Rg} are variations of random forests, algorithms labelled by \texttt{xgb} are variations of the boosting algorithm. Algorithms labelled with \texttt{SL} are implemented in the \texttt{SuperLearner} R package.   \par}
\end{minipage}
\end{figure}

\begin{table}[ht]
 \caption{Non-zero weights corresponding to the algorithms in the convex Super Learner, for estimating propensity scores.}
\label{pscoefs}
\centering
\begin{tabular}{rr}
  \hline
 & Algorithm's weight in ensemble \\ 
  \hline
  Random Forest (500 trees, 5 variables) & 0.18 \\ 
 Random Forest (2000 trees, 5 variables) & 0.36 \\ 
  Generalised additive models degree 3 & 0.46 \\ 
  GLM with all 2-way interactions & 0.01 \\ 
   \hline
\end{tabular}
\end{table}

\par 
For each propensity score and matching approach, we compare balance on the covariates in the data (reweighted by IPTW weights or by frequency weights from the matching, respectively).  Figure \ref{sdplot} displays the absolute standardised differences (ASD) for all the covariates, starting from the variables that were least imbalanced in the unweighted data, moving towards the more imbalanced. Generally, all weighting approaches tend to improve balance compared to the unweighted data, except for variables that were well balanced (ASD $< 0.05$) to begin with. Using the rule of thumb of $0.1$ as a metric of significant imbalance, we find that TWANG, when used for weighting, achieves acceptable balance on all covariates, except for the binary variable indicating having at least secondary education.  Based on the more stringent criterion of ASD $< 0.05$, however, TWANG leaves several covariates imbalanced, including the indicator of rural community, the availability of health center, the availability of birth clinic, and whether the mother can write in Indonesian.  When used for pair matching, the boosted CART based propensity score leaves high imbalances. This is expected as the balance metric in the loss function used the weighted, and not the matched data. The SL-based propensity score results in the largest imbalance, again reflecting that the loss function was set to maximimise the cross-validated MSE of the propensity score model, and not to maximise balance \footnote{We note that an extension of the SL that optimises balance has been proposed by \cite{Pirracchio2018}.}.

The estimated ATE results in a 10\%  increase in the probability of giving birth attended by a health professional, among those with contributory insurance (vs those without). With all the adjustment methods, this effect decreases, indicating an important role of adjusting for observed confounders. As expected,  method that reported the largest imbalances, IPTW SL, reports an ATE closest to the unadjusted estimate.

\begin{figure}
\caption{Covariate balance compared across balancing estimators for the ATE}
 	\centering
 	\includegraphics[scale=0.85]{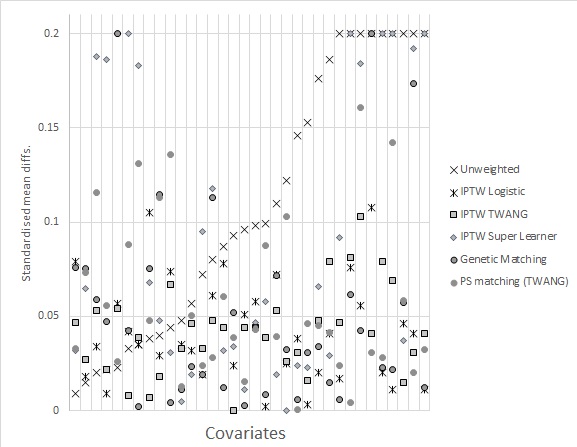}\label{sdplot}
\end{figure}
 	
\begin{table}[ht]
\centering
\label{balanceates}
\caption{ATEs and 95 \% CIs estimated using IPTW and matching methods}
\begin{tabular}{rrrr}
  \hline
 & ATE & 95 \% CI L & 95 \% CI U \\ 
  \hline
Unadjusted (naive) & 0.13 & 0.11 & 0.16 \\ 
  IPTW logistic & 0.06 & 0.02 & 0.11 \\ 
  IPTW TWANG & 0.08 & 0.04 & 0.12 \\ 
  IPTW SL & 0.11 & 0.09 & 0.13 \\ 
  PS matching TWANG & 0.08 & 0.04 & 0.12 \\ 
  Genetic Matching & 0.06 & 0.03 & 0.09 \\ 
   \hline
\end{tabular}
\end{table}

\subsection{Limitations of balancing methods}

Balancing methods allow for the consistent estimation of treatment effects, provided the assumptions of no unobserved confounding and positivity hold.  Crucially the analyst does not need to model outcomes thus increasing transparency  by avoiding cherry-picking of models. 
 While machine learning  can help by making the choices of the PS model or a distance metric data-adaptive, subjective choices remain. For example, the loss function needs to be specified, and if the loss function is based on balance, also the choice of balance metric. For example, the ASM chosen for demonstration purposes creates  a metric of average imbalance for the means. This ignores two potential complexities. First, imbalances in higher moments of the distribution are not taken into account by this metric.  While Kolmogorov-Smirnov tests statistics can take into account imbalance in the entire covariate  distribution \citep{stuart2010matching,diamond2013genetic}, and can be selected to enter the loss function for both the boosted CART and the Genetic Matching algorithms, there is a further issue remaining: how should the researcher trade-off imbalance across covariates?  With a large number of covariates, balancing one variable may decrease balance on an other covariate. Moreover, it is unclear how univariate balance measures should be summarised. The default of the TWANG package is to look at average covariate balance, while the Genetic Matching algorithm, as default, prioritises covariate balance on the variable that is the most imbalanced. This, however may not prioritise variables that are relatively strong predictors of the outcome, and any remaining imbalance would translate into a larger bias.  

Hence, there is an increasing consensus that exploiting information from the outcome regression generally improves on the properties of balance-based estimators \citep{kang2007demystifying,abadie2011bias}. ML methods to estimate the nuisance models for the outcome can provide reassurance against subjectively selecting outcome models that provide the most favourable treatment effect estimate. We review these methods in the following section.

\subsection{Machine learning methods for the outcome model}\label{OutReg}

Recall that under the unconfoundedness and positivity assumptions, the ATE can be identified by $E[Y|A=1,\mathbf{X}] - E[Y|A=0,\mathbf{X}]$, reducing the problem to one of estimation of these conditional expectations \citep{imbens2009recent,hill2011bayesian}. Denoting the true conditional expectation function for the observed outcome as 
$\mu(A,\mathbf{X}) = E[Y|A,\mathbf{X}]$, the regression estimator for the ATE can be obtained as 
\begin{equation}\label{reg}
\hat{\psi}^{reg} = \frac{1}{N}  \sum_{i=1}^{N} \big\{ \hat{\mu}(A=1,\mathbf{X}_i)-\hat{\mu}(A=0,\mathbf{X}_i) \big\},
\end{equation}
where $\hat{\mu}(A=a,\mathbf{X}=\mathbf{x})$  can
interpreted as the predicted potential outcome for level $a$ of the treatment among individuals with covariates $\mathbf{X} = \mathbf{x}$, and can be estimated by for example a regression 
$E[Y|A = a, \mathbf{X} = \mathbf{x}] =  \eta_0(\mathbf{x}) + \beta_1 a,$
with $\eta_0(\mathbf{x})$ the nuisance function and $\beta_1$ the parameter of interest for level $a$ of the treatment. 
 Under correct specification of the models for $\mu(A,\mathbf{X})$, the outcome regression  estimator is consistent \citep{bang2005doubly}, but it is prone to extrapolation.

The problem can now be viewed as a prediction problem, making it appealing to use ML to obtain good predictions for $\mu(A,\mathbf{X})$. Indeed, some  methods  do this: BARTs have been successfully used to obtain ATEs  \citep{hill2011bayesian,hahn2017bayesian}. \cite{austin2012using} demonstrates the use of a wide range of tree-based machine learning techniques to obtain regression estimators for the ATE. 

However, there are three reasons why ML is generally not recommended for outcome regression. First, the asympototic properties of such estimators are unknown. Typically, the convergence of the resulting regression estimator for the causal effect will be slower than $\sqrt{N}$ when using ML fits. A related problem is the so-called ``regularisation bias'' \citep{chernozhukov2017double,athey2018approximate}. Data-adaptive methods use regularisation to achieve optimal bias-variance trade-off, which  shrinks the estimates towards zero, introducing bias \citep{mullainathan2017machine}, especially if the shrunk coefficients correspond to  variable which are strong confounders \citep{athey2018approximate}. This problem increases as the number of parameters compared to sample size grows.   Third, it is difficult to conduct inference for causal parameters, as in general there is no way of constructing valid confidence intervals, and  the non-parametric bootstrap is not generally valid \citep{Bickel1997}.


This motivates going beyond single nuisance model ML plug-in estimators,  and using double-robust estimators with ML nuisance model fits, reviewed in the next section \citep{Farrell2015, chernozhukov2017double,athey2018approximate,Seaman2018}.

\subsection{Double-robust estimation with machine learning}\label{Sec:DR}

Methods that combine the strengths of outcome regression modelling with the balancing properties of the propensity score have been advocated for long. The intuition is that using propensity score matching or weighting as a ``pre-processing'' step can be followed by regression adjustment to control further for any residual confounding \citep{rubin2000combining,abadie2006large,imbens2009recent,stuart2010matching,abadie2011bias}. While these methods have performed well in simulations  \citep{kreif2013regression,busso2014new,kreif2016evaluating},  their asymptotic properties are not well understood.

A formal approach to combining outcome and treatment modelling was originally developed to improve the efficiency of IPTW estimators \citep{robins1995analysis}. \emph{Double-robust} (DR) estimators  use two nuisance models, and have the  special property that they are consistent as long as at least one of the two nuisance models is correctly specified. In addition, some DR estimators are shown to be semi-parametrically efficient, if both components are correctly specified \citep{robins2007comment}.
 
A simple DR method is the augmented inverse probability weighting ( AIPTW) estimator \citep{robins1994estimation}. The AIPTW can be written as 
$\psi^{AIPTW}=\psi^{AIPTW}(1)-\psi^{AIPTW}(0)$, where
\begin{equation}
\psi^{AIPTW}(a)
=\frac {1}{N}\sum _{i=1}^{N}\Biggl ({\frac {Y_{i}I({A_{i}=a})}{p(\mathbf{X}_i)}}- \frac {I({A_{i}=a})-p(\mathbf{X}_i)}{p(\mathbf{X}_i)}\mu(\mathbf{X}_i,A_i)\Biggr ),
\end{equation}
where $\mu(A,\mathbf{X}) = E[Y|A,\mathbf{X}]$, as before.

The variance of DR estimators is based on the variance of their \emph{influence function}.  
Let $\hat{\psi}$ be an estimator of a scalar parameter $\psi_0$, satisfying 
\begin{equation}
\sqrt{N}(\hat{\psi} -\psi_0)=\sqrt{N}^{-1}\sum_{i=1}^{N}\phi(O_i)+o(1),
\end{equation}
where $o(1)$ denotes a  term that converges in probability to 0, and where $E[\phi(O)]=0$ and $0<E[\phi(O)^2]<\infty$, i.e. $\phi(O)$ has zero mean and finite variance. Then $\phi(O)$ is the  \emph{influence function} (IF) of $\hat{\psi}$.  

By the central limit theorem,  the estimator $\hat{\psi}$ is asymptotically normal with asymptotic variance $N^{-1}$ times the variance of its influence function using this to construct normal-based confidence intervals.

A consequence of this convergence behaviour is that good asymptotic properties of DR estimators can be achieved even when the convergence rates of the nuisance models are slower than the conventional $\sqrt{N}$, as the DR estimator $\hat{\psi}$ can still converge at a fast $\sqrt{N}$ rate, as long as the product of the nuisance models convergence rates is faster than $\sqrt{N}$ (under regularity conditions, and empirical process conditions (e.g. Donsker class which can be avoided via sample splitting  \citep{Bickel/Kwon:2001, vanderLaanRobins2003,chernozhukov2017double}, described later). 

This discovery allows for the use of flexible machine learning-based estimation of the nuisance functions, leading to an increased applicability of DR estimators, which were previously criticised, given that most likely both nuisance models are  misspecified \citep{kang2007demystifying}. Concerns about the  sensitivity to  extreme propensity score weights remain \citep{petersen2012diagnosing}.

To improve on AIPTW estimators,
\citet{van2006targeted} introduced \emph{targeted minimum loss based estimation} (TMLE), a class of double-robust semiparametric efficient estimators.  TMLEs  ``target'' or de-bias an initial estimate of the parameter of interest in two stages  \citep{gruber2009targeted}. In the first stage, an initial estimate $\mu^0(A,\mathbf{X})$ of $E[Y|A,\textbf{X}]$ is obtained (typically by machine learning), and used to predict potential outcomes under both exposures, for each individual \citep{van2011targeted}.

In the second stage, these initial estimates are ``updated'', by fitting a generalised linear model for $E(Y|\mathbf{X})$, typically with logit link, an offset term $\mbox{logit}\{\mu^0(A,\mathbf{X})\}$ and a single so-called \emph{clever} covariate. When the outcome is continuous, but bounded, the update can also be performed on the logit scale \citep{gruber2010targeted}. For the ATE, the clever covariate is  $h(A,k) =\frac{A}{p(\mathbf{X})} - \frac{(1-A)}{(1-p(\mathbf{X}))}$.
The coefficient  $\epsilon$ corresponding to the clever covariate is then used to update (de-bias) the estimate of $\mu^0(A,\mathbf{X})$. The updating procedure continues until a step is reached where $\epsilon=0$. The final update $\mu^*(A,\mathbf{X})$ is the TMLE. For the special case of the ATE, convergence is mathematically guaranteed in one step, so there is no need to iterate. 

This exploits the information in the treatment assignment mechanism and ensures that the resulting estimator stays in the appropriate model space, i.e. it is a \emph{substitution estimator}. Again, data adaptive estimation of the propensity score is recommended \citep{van2011targeted}. Available software implementation of TMLE (R package \texttt{tmle} \citep{gruber2011tmle}) incorporates a Super Learner algorithm to provide the initial predictions of the potential outcomes and the propensity scores.

Another DR estimator with machine learning is the so-called \emph{double machine learning} (DML) estimator \citep{chernozhukov2017double}. For a simple setting of the ATE, this estimator simply  combines the residuals of an outcome regression and the residuals of a propensity score model, into a new regression, motivated by the partially linear regression approach of \cite{robinson1988root}. For the more general case when the treatment can have an interactive effect with covariates, the form of the estimator corresponds to the AIPTW estimator, where  the nuisance parameters are estimated using  machine learning algorithms. While the estimator does not claim ``double-robustness'' (as it does not aim to ``correctly specify'' any of the models), it aims to de-bias estimates of the average treatment effects by combining ``good enough'' estimates of the nuisance parameters. The machine learning methods used can be highly data-adaptive. This estimator is also semiparametric efficient, under weak conditions, due to an extra step of sample splitting (thus avoiding empirical process conditions \citep{Bickel/Kwon:2001}). The estimator is constructed by using  ``cross-fitting'', which divides the data into $K$ random splits, and witholds one part of data from fitting the nuisance parameters, while using the rest of the data to obtain predictions and constructing the ATE. This is then repeated $K$ times, and the average of the resulting estimates is the DML estimate for the ATE. The  standard errors are based on the influence function \citep{chernozhukov2017double}. Sample splitting is designed to help avoid overfitting, and thus reduces the bias.  A further adaptation of the method also takes into account the uncertainty about the particular sample splitting, by doing  large number of re-partitioning of the data, and taking the mean or median of the resulting estimates as the final ATE, and also correct the estimated standard errors to capture the spread of the estimates. 


\subsubsection{Demonstration of DR and double machine learning approaches using the birth data}
We begin by fitting a parametric AIPTW using logistic models for both PS and outcome regression. SEs are estimated by nonparametric bootstrap. We then use Super Learner (SL) fits for both nuisance models (with the  libraries described in Section 5.3). For the outcome models, we fit two  separate prediction models, for the treated and control observations, and obtain the predictions for the two expected potential outcomes, the probabilities of assisted birth under no health insurance and health insurance, given the individual's observed covariates. We plug these predictions into the standard AIPTW. The SEs are based on the influence function (without further modification).

Next, we implement the TMLE using the same nuisance model SL fits. SEs are based on the efficient influence function, as coded in the R package \texttt{tmle}. Finally, the double machine-learning estimates for the ATE are obtained using one-split of approximately equal size. The nuisance models are re-estimated in the first split of the sample using the SL with the same libraries as before, and obtaining predictions for the other half of the split sample. We then (in the cross-fitting step)  switch the roles of the two samples, and average the resulting estimates with the formulae for SEs based on the influence function, as in \citep{chernozhukov2017double}.

The relative weights of the candidate algorithms in the SL library are displayed in Table \ref{SLoutcome}, showing that the highly data-adaptive algorithms (boosting, random forests and BART) received the majority of the weights.  The estimated ATEs with 95\% CIs are reported in Table \ref{Table:DRres}. While the point estimates are of similar magnitude ($5-8$\% increase in the probability of assisted birth), their confidence intervals show a large variation. The SL AIPTW and TMLE appear to be very precisely estimated, displaying a narrower CI than the parametric AIPTW, where CIs have been obtained using parametric bootstrap. One potential explanation may that without the sample splitting, the nuisance parameters may be overfitted, and the influence function based standard errors do not take this into account. This is consistent with the finding of \cite{dorie2017automated} in simulated datasets that TMLE results in under-coverage of 95\% CIs. Indeed, the DML estimator, which only differs from the SL AIPTW estimator in the cross-fitting stage, displays the widest CIs, including zero. 

\begin{table}[ht]
\centering
\caption{Non-zero weights corresponding to the algorithms in the convex Super Learner, for modelling the outcome}
\label{SLoutcome}
\begin{tabular}{lcc}
  \hline
   & \multicolumn{2}{c}{Algorithm's weight in SL } \\ 
 & Model for control & Model for treated \\ 
  \hline
\small
 Boosting (100 trees, depth of 4, shrinkage 0.1) & 0.02 & 0.38 \\ 
 Boosting (1000 trees, depth of 4, shrinkage 0.1) & 0.00 & 0.04 \\ 
 Random forest (500 trees, 5 variables) & 0.00 & 0.36 \\ 
 Random forest (500 trees, 8 variables) & 0.00 & 0.09 \\ 
 Random forest (2000 trees, 8 variables) & 0.32 & 0.00 \\ 
 BART  & 0.54 & 0.00 \\ 
 GLM with no interaction  & 0.00 & 0.11 \\ 
 GLM with all 2-way interactions & 0.12 & 0.01 \\ 
   \hline
\end{tabular}
\end{table}

\begin{table}\centering
  \caption{ATE obtained with logistic AIPTW, Super Learner fits for the PS and outcome model AIPTW (SL AIPW), TMLE, and DML estimators applied to the birth data}\label{Table:DRres}
    \begin{tabular}{lrrr} \hline
          & \multicolumn{1}{l}{ATE} & \multicolumn{2}{c}{95\% CI}\\ \hline
   AIPTW (boot SE)  & 0.066 & 0.029 & 0.103 \\
   SL AIPTW  & 0.081 & 0.077 & 0.086\\
   TMLE & 0.073 & 0.065 & 0.081\\
   DML (1 split) & 0.053 & -0.026 & 0.133 \\
   \hline  
    \end{tabular}%
\end{table}%

\section{Variable selection}\label{Sec:VarSel}
A problem empirical researchers face when relying on a conditioning on a sufficient set of observed covariates for confounding control is variable selection, i.e. identifying which covariates  to include in the model(s) for conditional exchangeability to hold. In principle, subject matter knowledge should be used to select  a sufficient control set \citep{rubin2007design}. In practice however, there is often little prior knowledge
on which variables in a given data set are confounders. 
Hence data-adaptive procedures  to select the variables to adjust for 
become increasingly necessary when the number of potential confounders is very large. 
 There is a lack of clear guidance about what procedures to use, and about how to obtain valid inferences after variable 
selection. In this Section, we consider some approaches for variable selection when the focus is on the estimation of causal effects. 

Decisions on whether to include a covariate in a regression model, whether these are done manually or by automated methods, such as stepwise regression, are usually  based on the
strength of evidence for the residual association with the outcome, by for example, iteratively testing for significance in models that include or exclude the variable, and comparing the resulting p-value to a pre-specified significance level. Stepwise models (backwards or forwards selection) are however widely recognised to perform poorly \citep{Heinze2018}, for two main reasons.  First, collinearity can be an issue, which is especially problematic for forward-selection, while in high-dimensional settings backward selection may be unfeasible. Second, tests performed during the variable selection process are not pre-specified, and this is typically not acknowledged  in the subsequent analysis, compromising the validity and the interpretability of subsequent inferences, derived  from models after  variable selection.

Decisions about which covariates to adjust for in a regression must ideally be based on the evidence of confounding, taking into account the covariate-exposure association. Yet causal inference procedures that only rely on the strength of covariate-treatment relationships (e.g. propensity score methods) may also be problematic. For example, they may lead to adjusting for variables that are causes of the exposure only (so-called pure instruments), inducing bias \citep{Vansteelandt2012}. On the other hand, if variable selection only relies on modelling the outcome, using  for example, lasso regression, it may introduce regularisation bias, due to underestimating coefficients, and as a result, mistakenly excluding variables with non-zero coefficients. 

To address these challenges, \citet{Belloni2014} proposed a solution that offers principled variable selection, taking into account both the covariate-outcome and the covariate-treatment assignment association, resulting in valid inferences after variable selection.  Their framework, referred to as ``post double selection'', or ``double-lasso', also allows to extend to space of possible confounding variables to  include higher order terms. Following \citet{Belloni2014}, we consider the partially linear model $Y_i=g(\mathbf{X}_i)+ \beta_0 A_i + \zeta_i$, where $\mathbf{X}_i$ a set of confounder-control variables, and $\zeta_i$ is the error term satisfying $E[\zeta_i |A_i,\mathbf{X}_i]=0$.
We examine the problem of selecting a set of variables $\mathbf{V}$
from among $d_2$ potential variables $W_i=f(\mathbf{X})$, which includes $\mathbf{X}$ and transformations of $\mathbf{X}$ as  to adequately approximate $g(\mathbf{X})$, and allowing for $d_2>N$. Crucially, pure instruments, i.e. variables associated with the treatment but not the outcome, do not need to be identified and excluded in advance.

We identify covariates for inclusion in our estimators of causal effects in two steps. First we find those that predict the outcome and in a separate second step those that predict the treatment. A lasso linear regression calibrated to avoid overfitting is used for both models. In a final step, the union of the variables selected in either step is used as the confounder control set, to be used in the causal parameter estimator. The control set can include some additional covariates identified beforehand. 

\citet{Belloni2014} show that the double lasso  results in valid estimates of ATE under the key  assumption of \textit{ultra sparsity}, i.e. conditional exchangeability holds after controlling for a relatively small number $s<<\sqrt{N}$  of variables in $\mathbf{X}$ not known apriori. 
Implementation is straightforward, for example using the \texttt{glmnet} R package. We use cross-validation for choosing  the tuning parameter, following \citet{dukes2018high}. Once the confounder control set is selected, a standard method of estimation is used, for example ordinary least squares estimation of the outcome regression.

\subsection{Application of double lasso to the birth data}

We now apply the double lasso approach for variable selection to our birth data example.  We begin by running a lasso linear regression (using the \texttt{glmnet} R package) for both outcome and treatment separately, including all the variables available and using cross-validation to select the penalty.

The union of the variables selected for both models was all 35 available covariates. These variables are used to control for confounding first in a parametric logistic outcome regression model, which we use to predict the potential outcomes, and obtain the ATE. We also calculate IPTW and AIPTW estimates using the weights from a parametric logistic  model for the PS. For all estimates we use bootstrap to obtain SEs.

We then increase the covariate space to include all the two-way interactions between the covariates, excluding the exposure and the outcome, resulting in a total of 595 covariates. Using double-lasso on this extended covariate space, we select 156 covariates for the outcome and  89 for the treatment model, leaving us a union set used for confounding control of 211. We repeat the three estimators now based on this expanded control set.

Table \ref{tab:doublelasso} reports the esimated ATEs and 95 \% CIs. The CI's for the double-lasso outcome regression  were obtained using the non-parametric bootstrap, while the IPTW and AIPTW were obtained as before using the sandwich SEs.
The top panel of the table shows estimates using all covariates but no interactions, and the bottom panel shows estimates using 72 covariates, including the main terms and the interactions. The point estimates change very little, implying a minor role of the interactions in controlling for confounding.  

\begin{table}[h]
  \centering
  \caption{ATE post-double-lasso for selection of confounders applied to the birth data} \label{tab:doublelasso}%
    \begin{tabular}{lrrr} \hline
          & \multicolumn{1}{l}{ATE} & \multicolumn{2}{c}{95\% CI}\\ \hline
    Outcome Regression   & 0.027&  0.012 &  0.041 \\
    IPTW   & 0.083 & 0.016 & 0.107 \\
    AIPTW  & 0.066 & 0.029 & 0.103 \\
    \hline
    with 2-way interactions \\
    \hline
    Outcome Regression & 0.032  & 0.016 & 0.046 \\
    IPTW & 0.078 & 0.019 & 0.107 \\
    AIPTW  & 0.063 & 0.026 & 0.101 \\
     \hline
    \end{tabular}%
 \end{table}%

\subsection{Collaborative TMLE}

 Covariate selection for the propensity score can also be done within the TMLE framework.
Previously, we have seen that for a standard TMLE estimator of the ATE, the estimation of the propensity score model is performed independently from the estimation of the initial estimator of the outcome model, i.e. without seeking to optimise the fit of the PS for the estimation of the target parameter.  

However, it is possible, and even desirable, to choose the treatment model  which is optimised to reduce the mean square error the target parameter. An extension of the TMLE framework,  the so-called collaborative TMLE (CTMLE) does just this. 

The original version of CTMLE  \citep{VanDerLaan2010} is often referred to as  ``greedy CTMLE''.  
Here, a sequence of nested logistic regression PS models is created by a greedy forward stepwise selection algorithm, nested 
such that at each stage (i.e. among all PS models with $k$ main terms), we select the PS model which results in the estimated MSE of the ATE parameter being the smallest, when used in the targeting step to update the initial estimator of the outcome model. If the resulting TMLE does not improve upon the empirical fit of the initial outcome regression, then this TMLE is not included in the sequence of TMLEs. Instead, the initial outcome regression estimator is replaced by the last previously accepted TMLE and we start over. This procedure is performed iteratively until all $d$ covariates have been incorporated into the PS model.
The greedy nature of the algorithm  makes it computationally intensive: the total number of models explored is $d + (d-1) + (d-2)+ ... + 1$, which is of the order $O(d^2)$ \citep{Ju2017}. 

This has led to the development of scalable versions \citep{Ju2017} of CTMLE, which replace the greedy search with a data-adaptive pre-ordering of the candidate PS model estimators. The time burden of these scalable CTMLE algorithms is of order $O(d)$. Two CTMLE pre-ordering strategies are proposed: \emph{logistic} and \emph{correlation}.
The logistic preordering CTMLE constructs an univariable estimator of the PS for each available covariate, $p_{k}$ with $X_k$ the baseline variable, $k=1,\ldots,d$. Using the resulting predicted PS,  we construct the clever covariate corresponding to the TMLE for the ATE, namely $h(A,k) =\frac{A}{p_{k}} - \frac{(1-A)}{(1-p_{k})}$, 
and fluctuate the initial estimate of the parameter of interest $\mu^0$ using this clever covariate, (and a logistic log-likelihood) as usual in the TMLE literature. We obtain the targeted estimate and compute the empirical loss, which could be for instance the mean sum of squares errors. Finally we order  the covariates by increasing value of this loss. 

The correlation pre-ordering is motivated by noting that we would like the $k$-th covariate added to the PS model to be that which best explains the current residual, i.e. between $Y$ and current targeted estimate. Thus, the correlation pre-ordering ranks the covariates based on their correlation with the residual between $Y$ and the initial outcome regression estimate $\mu^0$ \citep{Ju2017}.

For both pre-orderings,  at each step of the CTMLE, we add the variables to the PS prediction model in this order, as long as the value of the loss function continues to decrease. 

Another version of the CTMLE exploits lasso regression  for the selection of the variables to be included in the PS estimation \citep{Ju2017b}.  This CTMLE algorithm also constructs a sequence of propensity score estimators, each of them a lasso logistic model with a penalty $\lambda_k$, where $k$ is monotonically decreasing, and which is ``initialised'' with $\lambda_1$ the minimum $\lambda$ selected by cross-validation.
Then, the corresponding TMLE estimator for the ATE is constructed for each PS model,  finally choosing by cross-validation the TMLE which minimises the loss function (MSE). The SEs for all  CTMLE versions  are computed based on the variance of the influence function, as implemented in the R package \texttt{ctmle}.

\subsubsection{CTMLE applied to the birth data}
We now apply all these variants of CTMLE to the case study. The final logisic pre-ordering CTMLE is based on a PS model containing nine covariates, including variables indicating the year of birth, and variables capturing education and socioeconomic status.
The CTMLE based on correlation pre-ordering selected three covariates for the PS, one variable that captures participation in a social assistance programme, but also two variables that measure the availability of health care providers in the community. These latter variables are indeed to be expected to have a strong associaton with the outcome, assisted birth, hence it is not surprising that they were selected, based on their role in reducing the MSE of the ATE.
Finally, the lasso CTMLE is based on a penalty term 0.000098 chosen by cross-validation, which resulted in all variables having non-zero coefficients for the PS.   

The results are reported on Table \ref{Table:CTMLEres}. The estimated ATEs are somewhat larger than those obtained using TMLE, all reporting an increase of around 8\% in the probability of giving birth assisted by a health care professional, for those who have social health insurance vs. uninsured.  However, the CTMLE estimate resulting from the lasso, which selected all variables into the PS model, is (unsurprisingly) very similar to the TMLE estimate reported in Section \ref{Sec:DR}, and thus is further away from the naive estimate than the estimates from CTMLEs that use  only a subset of the variables. Lasso-CTMLE  also has wider 95\% CIs than the rest, while the greedy CTMLE has the tightest. This may be an empirical evidence of the bias-variance trade off:  due to using less covariates in the PS, the estimators that use aggressive variable selection for the PS are slightly biased, but with lower variance.

\begin{table}[h]
  \centering
  \caption{ATE obtained with CTML estimators applied to the birth data}\label{Table:CTMLEres}
    \begin{tabular}{lrrr} \hline
          & \multicolumn{1}{l}{ATE} & \multicolumn{2}{c}{95\% CI}\\ \hline
    Greedy  CTMLE & 0.082   &(0.071, 0.094) \\
      Scalable   CTMLE logistic pre-ordering &  0.085 & (0.067, 0.104) \\
    Scalable  CTMLE correlation pre-ordering   &0.082 &   (0.068, 0.097) \\
 Lasso CTMLE & 0.076 & (0.048, 0.105) \\
 \hline  
    \end{tabular}%
\end{table}%

\section{Further topics}
Throughout this chapter, we have seen how machine learning can be used in estimating ATE, by using it as a prediction tool for outcome regression or PS models. The same logic can be applied to many other estimation problems where there are nuisance parameters that need to be predicted as part of the estimation of the parameter of interest.

\subsection{Estimating treatment effect heterogeneity}
We focused the discussion to the common target parameter, the ATE. Most of the methods considered are also available for the ATT parameter (see e.g. \cite{chernozhukov2016double}). The difference between the ATE and ATT stems from heterogeneous treatment effects, and this heterogeneity -- in particular, heterogeneity with respect to observed covariates --  in itself can be an interesting target of causal inference.  For example, in the birth cohort example of this paper, we may be interested in how the effect of health insurance varies by the socioeconomic gradient.  To answer this question, one may either want to specify the ``treatment effect function'', a possibly non-parametric function of the treatment effects as a function of deprivation, and possibly other covariates such as age. An other approach may be subgroup analysis, based on variables that have been selected in a data-adaptive way. 

\citet{imai2013estimating} propose a variable selection method using Support Vector Machines, to estimate heterogeneous treatment effects in randomised trials.  \citet{hahn2017bayesian} further develop the BART framework to estimate treatment effect heterogeneity, by flexibly using the estimated propensity score to correct for regularisation bias. 
\citet{athey2016recursive} propose a regression tree method, referred to as ``causal trees'' to identify subgroups with treatment effect heterogeneity, using sample-splitting to avoid overfitting. \citet{wager2017estimation} extend this idea to random forest-based algorithms, referred to as ``causal forests'' for which they establish theoretical properties.  
 A second interesting question may concern optimal policies: if a health policy maker has limited resources to provide free insurance for those currently uninsured, what would be an optimal allocation mechanism that maximimses health benefits? The literature on ``optimal policy learning'' is rapidly growing. \citet{kitagawa2017should} focus on estimating optimal policies from a set of possible policies with limited complexity, while \cite{athey2017efficient} further develop the double-machine learning \citep{chernozhukov2016double} to estimate optimal policies.  Further approaches have been proposed, for example \citep{kallus2017balanced} based on balancing and within the TMLE framework  \citep{van2015targeted,luedtke2017faster}.

\subsection{Instrumental variables} \label{sec:IV}

In certain situations, even after adjusting for observed covariates, there may be doubts as to whether conditional exchageability holds. However, other methods can be used where there is an instrumental variable (IV) available, that is a variable which is correlated with the exposure but is not associated with any confounder of the exposure-outcome association, nor is there any pathway by which the IV affects the outcome, other than through the exposure.  
Depending on the additional assumptions the analyst is prepared to make,  different estimands can be identified. Here, we focus on the local average treatment effect (LATE) under monotonic treatment \citep{Angrist1996}. 

Consider the (partially) linear instrumental variable model,  which in its simpler form can be thought of as a two-stage procedure, where the first stage consists of a linear regression of the endogeneous exposure $A$ on the instrument $Z$, $A = \alpha_0+ \alpha Z + \epsilon_a$ . Then in a second stage, we regress the outcome on the predicted exposure $\hat{A}$, $Y=\beta_0 + \beta_1 \hat{A} + \epsilon_y$. 
 
 Usually, the first stage is treated as an estimation step, and the coefficients are obtained using OLS. In fact,  we are only interested in the predicted exposure for each observation, and the parameters in the first stage are merely nuisance parameters that must be estimated to calculate the  fitted values for exposure. Thus, we can think of this problem directly as a prediction problem, and use machine learning algorithms for the first stage. This can help alleviate some of the finite sample bias, often observed in IV estimates, which are typically biased towards the OLS, as a consequence of overfitting the first stage regression, a problem that is more serious with small sample sizes or weak instruments \citep{mullainathan2017machine}.

A number of recent studies have used ML for the first stage of the IV models. 
\citet{Belloni2012} use lasso, while \citet{Hansen2014} use ridge regression. More recently, a TMLE has been developed for IV models, which uses ML fits also for the initial target parameter estimation \citep{Toth2016}. Double robust IV estimators can also be used with machine learning predictions for the nuisance models in the second stage, as shown in \citep{chernozhukov2016double, DiazOrdaz2018}.

\section{Discussion}

We have attempted to provide an overview of the current use of ML methods for causal inference, in the setting of evaluating the average treatment effects of binary static treatments, assuming no unobserved confounding.  We used a case study of an evaluation of a health insurance scheme on health care utilisation in Indonesia. The case study displayed characteristics typical of applied evaluations: a binary outcome, and a mixture of binary, categorical and continuous covariates. A practical limitation of the presented case study is the presence of missing data. While for simplicity,  we used a complete case analysis, assuming that missingness does not depend on both the treatment and the outcome \citep{Bartlett2015}. If this is not the case, the resulting estimates may be biased. Several options to handle missing data under less restrictive assumptions exist: e.g. multiple imputation \citep{Rubin:1987disc}, which is in general valid under missing-at-random assumptions, or the missing indicator method, which includes indicators for missingness in the set of potential variables to adjust for, and relies on the assumption that the confounders are only such when observed \citep{DAgostino2001}. Another alternative is to use inverse probability of ``being a complete case'' weights, which can be easily combined with many of the methods described in this article \citep{Seaman2014}.

We have highlighted the limitations of naively interpreting the output of machine learning prediction methods as causal estimates, and provided a review of recent innovations that plug-in ML prediction of nuisance parameters in ATE estimators. We have demonstrated how ML can make the estimation of the PS more principled, and also illustrated a multivariate matching approach that uses ML to data-adaptively select balanced comparison groups. We also highlighted the limitations of such ``design based'' approaches: they may not improve balance on variables that really matter to reduce bias in the estimated ATE, as they  cannot easily  take into account information on the relative importance of confounders for the outcome variable.

We gave a brief overview of the possibility of using ML for estimating ATEs via outcome regressions. We emphasised that obtaining valid confidence intervals after such procedures is complicated, and the bootstrap is not valid. Some methods, such as BARTs  are able to give inferences based on the corresponding posterior distributions, and have been used in practice with success \citep{dorie2017automated}. Nevertheless,  there are currently no theoretical results underpinning its use \citep{wager2017estimation}, and thus BART inferences should be used with caution. Instead, we illustrated double-robust approaches that combine the strengths of PS estimation and outcome modelling, and are able to incorporate ML predictors in a principled way. These approaches, specifically TMLE pioneered by van der Laan and colleagues, and the double machine learning estimators developed by Chernozokov and colleagues, have appealing theoretical properties and increasing evidence of their good finite sample performance \citep{porter2011relative,dorie2017automated}. 

All estimation approaches demonstrated in this article rely on the assumption that selection into treatment is based on observable covariates only. In many settings of policy evaluations, this assumption is not tenable. Under such settings, beyond instrumental variable methods (discussed in Section \ref{sec:IV}), panel data approaches are commonly used to control for one source of unobserved confounding, that is due to unobservables that remain constant over time. To date, ML approaches have not been combined with panel data econometric methods. Exceptions are \cite{bajari2015machine} and \cite{chernozhukov2017orthogonal} who demonstrate ML approaches for demand estimation using panel data.

We stress once again that ML methods can improve the estimation of causal effects only once the identification step has been firmed up and using estimators with appropriate convergence rates, so that they remain consistent even when using ML fits. However, with the increasing availability of Big data, in particular in settings with a very large number of covariates, assumptions such as ``no unobserved confounders'' may be more plausible \citep{titiunik2015can}. With such $d>>n$ datasets, ML methods are indispensable for variable selection as well as the construction of low dimensional parameters such as average treatment effects. Indeed, many innovations in ML for causal inference are taking place in such $d>>n$ settings (e.g. \citet{belloni2011L1,belloni2014inference,wager2017estimation}).

Finally, we believe that, paradoxically, ML methods that are often criticized for their `'black box'' nature, may increase the transparency of applied research. In particular, ensemble learning algorithms such as the Super Learner, can provide a safeguard against having to hand-pick the best model or algorithm. ML should be encouraged to enhance expert substantive knowledge when selecting confounders and model specification.

\section*{Acknowledgements}

Karla DiazOrdaz was supported by UK Wellcome Trust Institutional Strategic Support Fund-- LSHTM Fellowship 204928/Z/16/Z.

No\'emi Kreif gratefully acknowledges her co-authors on the impact evaluation of the Indonesian public health insurance  scheme: Andrew Mirelman, Rodrigo Moreno-Serra, Marc Suhrcke (Centre for Health Economics, University of York) and Budi Hidayat (University of Indonesia).

\def\refname{References}

\bibliography{bibl}

\end{flushleft}

\end{document}